\documentclass[runningheads]{llncs}
\usepackage[T1]{fontenc}

\usepackage{tikz}
\usepackage{pifont}
\newcommand{\cmark}{\ding{51}}%
\newcommand{\xmark}{\ding{55}}%
\usepackage{xcolor} 
\usepackage{amssymb} 
\usepackage{graphicx}
\usepackage{latexsym}
\usepackage{amsmath}
\usepackage{hyperref}
\begin{document}
\title{VMambaMorph: a Multi-Modality Deformable Image Registration Framework based on Visual State Space Model with Cross-Scan Module}
\titlerunning{VMambaMorph}

\author{Ziyang Wang\inst{1} \and Jian-Qing Zheng\inst{1} \and
Chao Ma\inst{2} \and Tao Guo }
\authorrunning{Z. Wang et al.}
%
\institute{University of Oxford, UK \\
\email{ziyang.wang@cs.ox.ac.uk}\\ 
\and
Mianyang Visual Object Detection and Recognition Engineering Center, China
}
\maketitle              

\begin{abstract}
Image registration, a critical process in medical imaging, involves aligning different sets of medical imaging data into a single unified coordinate system. Deep learning networks, such as the Convolutional Neural Network (CNN)-based VoxelMorph, Vision Transformer (ViT)-based TransMorph, and State Space Model (SSM)-based MambaMorph, have demonstrated effective performance in this domain. The recent Visual State Space Model (VMamba), which incorporates a cross-scan module with SSM, has exhibited promising improvements in modeling global-range dependencies with efficient computational cost in computer vision tasks. This paper hereby introduces an exploration of VMamba with image registration, named VMambaMorph. This novel hybrid VMamba-CNN network is designed specifically for 3D image registration. Utilizing a U-shaped network architecture, VMambaMorph computes the deformation field based on target and source volumes. The VMamba-based block with 2D cross-scan module is redesigned for 3D volumetric feature processing. To overcome the complex motion and structure on multi-modality images, we further propose a fine-tune recursive registration framework. We validate VMambaMorph using a public benchmark brain MR-CT registration dataset, comparing its performance against current state-of-the-art methods. The results indicate that VMambaMorph achieves competitive registration quality. The code for VMambaMorph with all baseline methods is available on GitHub. 

\url{https://github.com/ziyangwang007/VMambaMorph}

\end{abstract}
\section{Introduction}

Predicting for non-rigid anatomical shifts, deformable image registration stands as a pivotal component of medical image analysis \cite{sotiras2013deformable}. This vital process ensures the spatial alignment of images prior to their examination. Registration is instrumental in medical imaging, enabling the juxtaposition of images obtained over time (for longitudinal studies), from varied scanner technologies (multi-modal registration), or among different individuals for broader research studies (inter-subject registration).

The essence of image registration lies in deducing the spatial transformation, ${\phi}:\mathbb{R}^{n}\to\mathbb{R}^{n}$, which is articulated through parameters that define either a linear spatial transformation (like rigid or affine transformations) or a continuum of movements (or displacements) indicative of deformation, represented by $\phi[\textbf{\textit{x}}]\in \mathbb{R}^{d}$ at a specific coordinate $\textbf{\textit{x}}\in\mathbb{Z}^{d}$ within a target image $\textbf{\textit{I}}^{\rm tgt}\in\mathbb{R}^{{n}}$, originating from a source image $\textbf{\textit{I}}^{\rm src}\in\mathbb{R}^{n}$. Here, $n$ denotes the dimensions of a 3D image as $n={H \times W\times T}$, with $d, T, H, W$ corresponding to the dimensionality, thickness, height, and width of the image, respectively.
Traditionally approached as an optimization challenge, image registration endeavors to minimize a dissimilarity metric $\mathcal{D}$ alongside a regularization term $\mathcal{R}$:
\begin{equation}
\label{equ:opt_ddf}
\hat{\phi}=\underset{\phi}{\mathrm{argmin}}{\big(\mathcal{L}_{\rm dissim}(\phi(\textbf{\textit{I}}^{\rm src}),~\textbf{\textit{I}}^{\rm tgt})+\lambda\mathcal{L}_{\rm smooth}(\phi)\big)}
\end{equation}
In this formulation, $\phi$ signifies the hypothesized spatial transformation, while $\lambda$ is the regularization term's weight.

\begin{figure*}[ht!]  
\centering  
\includegraphics[width=\linewidth]{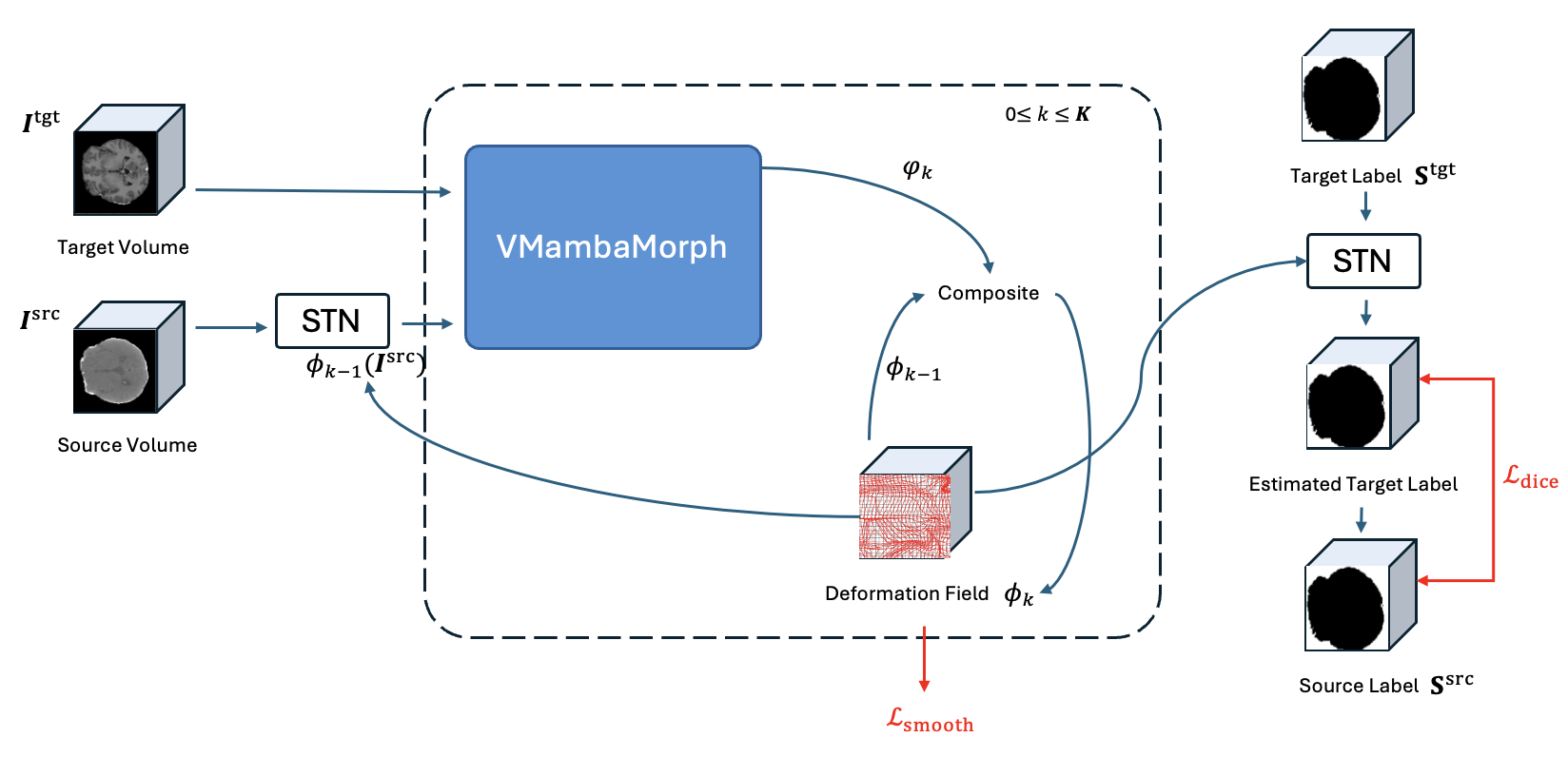}  
\caption{The Recursive Registration Framework of VMambaMorph.}  
\label{fig:recursive}  
\end{figure*}

While traditional registration techniques are capable of calculating diffeomorphic displacement fields with considerable accuracy, their substantial computational requirements and extended processing durations hinder their practicality for real-time applications. Over the last decade, registration methods powered by deep learning \cite{balakrishnan2019voxelmorph} have risen as efficient alternatives. These techniques deliver rapid registration capabilities and, in some instances, achieve accuracy comparable to that of conventional methods:
\begin{equation}
\label{equ:learn_ddf}
{\phi}=\mathcal{R}(\textbf{\textit{I}}^{\rm src},\textbf{\textit{I}}^{\rm tgt};{\textbf{\textit{w}}})\
\end{equation}
Here, $\mathcal{R}$ represents the deep learning model. This model is refined through a training process aimed at minimizing the loss function, exemplified by Equation~\eqref{equ:opt_ddf}.

In a typical learning-based framework, the source and target images/volumes are concatenated and fed into a single-branch network volume, which is then processed through a UNet-like architecture to compute the deformation field. 

Voxelmorph \cite{balakrishnan2019voxelmorph} is an early deep learning-based deformable image registration, using classic convolutional network structure, UNet \cite{ronneberger2015u}.
However, the convolutional network structure suffers from a limited receptive field, thus causing a limited capture range of deformation between two images.

The attention mechanism, as proposed by Vaswani et al. \cite{vaswani2017attention}, is designed to capture long-range dependencies in contextual sequential features. It has since been adapted for computer vision in the form of the Vision Transformer \cite{dosovitskiy2020image}, achieving a global receptive field capable of addressing the issue of limited capture ranges. Consequently, it has been employed in multiple studies for both rigid image registration \cite{zheng2020d,zheng2023accurate} and deformable image registration \cite{zhang2021learning,song2021cross,zheng2022recursive}. However, the attention mechanism is also known to incur high computational complexity. Several previous efforts \cite{kang2022dual,zheng2024residual} have investigated progressively coarse-to-fine image registration strategies to reduce computational demands while retaining the benefits of extensive capture range for deformation. Nevertheless, these methods also present challenges in preserving discontinuity in the predicted deformation \cite{zheng2024residual}.

In this landscape, the emerging Mamba framework, named SSM, which is designed for efficient feature processing in sequential features, offers a compelling alternative \cite{gu2023modeling,gu2023mamba,gu2021efficiently}. The initial iterations of Mamba have been succeeded by Visual Mamba \cite{liu2024vmamba}, which incorporates a selective scan module within the Mamba architecture, enhancing its performance in computer vision tasks. This advancement has led to a slew of Mamba-based networks being proposed for a variety of applications \cite{ma2024u,xing2024segmamba,wang2024mamba,wang2024semi,zhu2024vision,wang2024weak}. MambaMorph \cite{guo2024mambamorph} is one such innovation, utilizing a hybrid Mamba-CNN network as the registration module to compute deformation fields. Despite these developments, there remains an untapped potential in the latest Mamba techniques, especially considering the differences between sequential and visual features.

In this paper, we explore the application of Visual Mamba \cite{liu2024vmamba} to deformable image registration, which has been further developed to efficiently extract visual features from images and volumes. The contributions of this research are outlined as follows: 
\begin{enumerate}
\item To the best of our knowledge, this represents the first investigation into the use of the Visual State Space Model (VMamba) \cite{liu2024vmamba} for medical image registration tasks.
\item Inspired by the recent success of VMamba \cite{liu2024vmamba}, the 2D image-based Visual State Space (VSS) block is redesigned for 3D volumetric feature processing.
\item VSS blocks are integrated with a conventional CNN-based U-shaped network to serve as the registration module.
\item To overcome the challenge of complex motion and structure in multi-modality image registration, a recursive registration framework is adopted. In addition, a fine-grained feature extractor is utilized before registration.
\item The proposed registration method, VMambaMorph, is validated on a public benchmark MR-CT dataset, achieving performance that surpasses existing state-of-the-art networks for deformable image registration.
\end{enumerate}

\begin{figure*}[ht!]  
\centering  
\includegraphics[width=\linewidth]{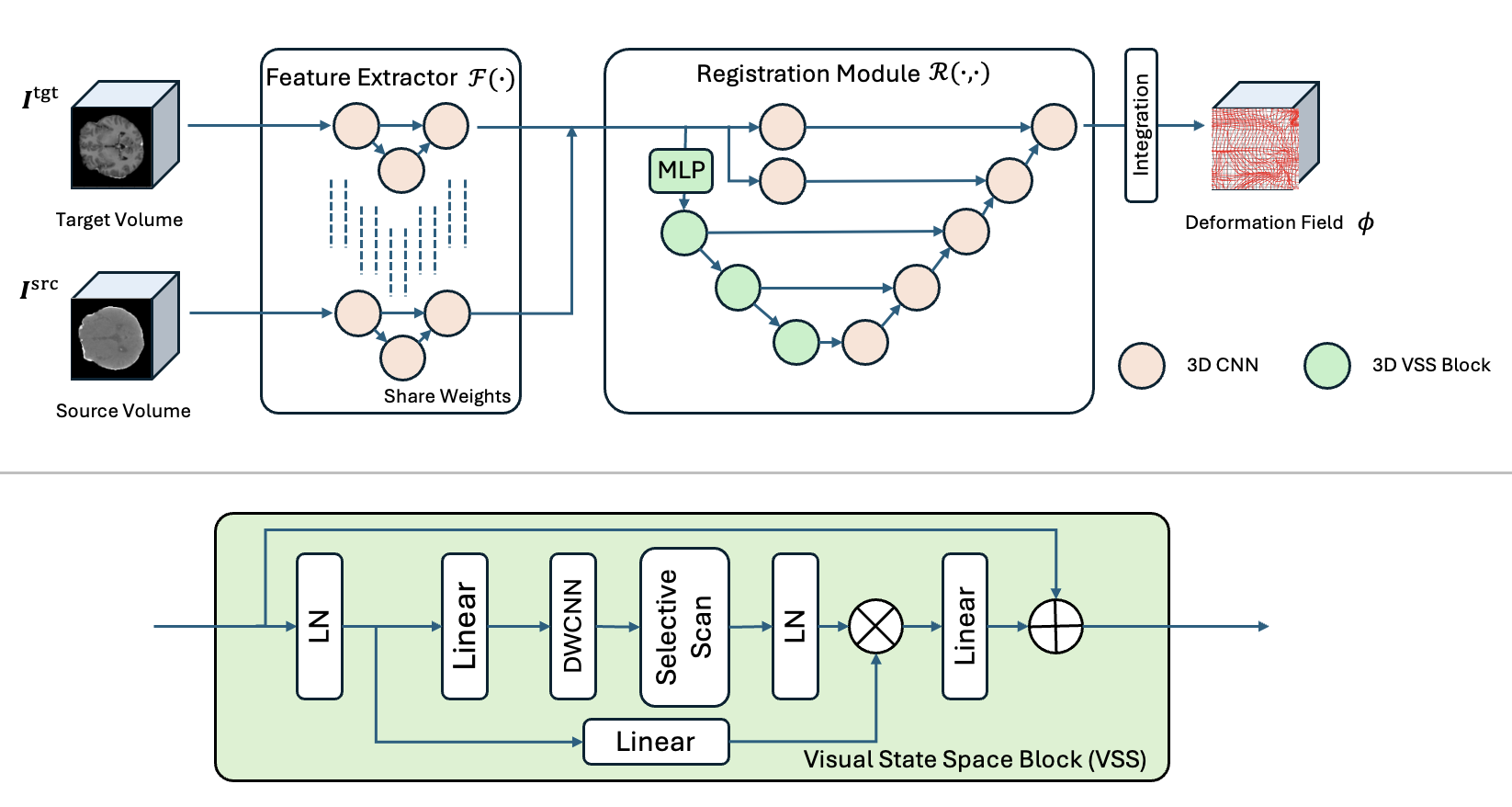}  
\caption{The Architecture of VMambaMorph, and the Details of Visual State Space Block.}  
\label{fig:framework}  
\end{figure*}

\section{Method}
The proposed VMamaMorph-based recursive registration framework is sketched in Figure~\ref{fig:recursive} and the network structure of VMambaMorph is shown in \ref{fig:framework}, consisting of a two-branch feature extractor module, registration module, and STN. A pair of target volume and source volume are respectively denoted as $\textbf{\textit{I}}^{\rm tgt}$ and $\textbf{\textit{I}}^{\rm src}$, and their corresponding segmentation is denoted as $\textbf{\textit{S}}^{\rm tgt}$ and $\textbf{\textit{S}}^{\rm src}$. The registration module is $\mathcal{R}(\cdot,\cdot)$ and the feature extractor is as $\mathcal{F}(\cdot)$. The deformation field is denoted as $\phi$. The losses include $\mathcal{L}_{\rm dice}(\cdot,\cdot)$ and $\mathcal{L}_{\rm smooth}(\cdot)$ following \cite{balakrishnan2019voxelmorph,guo2024mambamorph}.

\subsection{Visual State Space Model}

The CNN-based networks have been established as robust mechanisms for local feature extraction \cite{long2015fully,ronneberger2015u,balakrishnan2019voxelmorph,zhang2020novel}, and ViT-based networks are strength in global-range dependency modeling \cite{shi2022xmorpher,cao2022swin,wang2023dual,chen2022transmorph,zhang2021learning}. Although hybrids of CNN and ViT have been explored \cite{chen2021transunet,chen2022transmorph,wang2024mixsegnet}, they still incur significant computational costs due to ViT’s quadratic complexity. In this landscape, the emerging Mamba framework, named SSM, offers a compelling alternative. The initial iterations of Mamba have been succeeded by Visual Mamba \cite{liu2024vmamba}, which incorporates a selective scan module within the Mamba architecture, enhancing its performance in computer vision tasks. The conventional SSM is developed as a linear time-invariant system function to map $x(t) \in \mathbb{R} \mapsto y(t) \in \mathbb{R}$ through a hidden state $h(t) \in \mathbb{R}^N$, given $A \in \mathbb{C}^{N \times N}$ as the evolution parameter, $B, C \in \mathbb{C}^{N}$ as the projection parameters for a state size $N$, and skip connection $D \in \mathbb{C}^{1}$. The process is illustrated in Equation \ref{ode} known as linear ordinary differential equations.

\begin{equation}
\label{ode}
h'(t) = Ah(t) + Bx(t), y(t) = Ch(t) + Dx(t)
\end{equation}

Considering the input $x_{k} \in \mathbb{R}^{L \times D}$ a sampled vector within the signal flow of length $L$, the discrete version of this linear model can be transformed by zeroth-order hold rule given a timescale parameter $\Delta \in \mathbb{R}^{D}$ \cite{gu2021combining,gupta2022diagonal}.

\begin{equation}
h_t = \overline{A}h_{k-2} + \overline{B}x_k 
\end{equation}

\begin{equation}
y_t = {C}h_k + \overline{D}x_k
\end{equation}

\begin{equation}
\overline{A} = e^{\Delta A} 
\end{equation}

\begin{equation}
\overline{B} = (e^{\Delta A} - I) A^{-1}B 
\end{equation}

\begin{equation}
\overline{C} = C 
\end{equation}
where $B,C \in \mathbb{R}^{D \times N}$. The approximation of $\overline{B}$ refined using first-order Taylor series $\overline{B} = \left(e^{\Delta A} - I\right) A^{-1} B \approx \left(\Delta A\right)\left(\Delta A\right)^{-1} \Delta B = \Delta B$. The Visual Mamba further introduce Cross-Scan Module then integrate convolutional operations into the VSS block, which is detailed in \cite{gu2023mamba,liu2024vmamba}. We have redesigned the original VSS network blocks for the encoder of registration module for 3D image processing considering the volumetric medical image registration.

\subsection{Recursive Registration Framework}

Addressing the intricate challenge presented by complex motion and structural variations in multi-modality image registration, we adopt a recursive registration framework structure aimed at achieving coarse-to-fine registration, as depicted in Fig.~\ref{fig:recursive}. This initiative is inspired by \cite{zhao2019recursive,zheng2022recursive}. In our proposed method, the residual transformation $\varphi_{k}$, representing the alignment discrepancy between the target image $\textbf{\textit{I}}^{\rm tgt}$ and the successively warped source feature map $\phi_{k-1}(\textbf{\textit{I}}^{\rm src})$ from the previous $k-1$ registration level, is computed by the operator $\mathcal{R}$ and progressively refined through compositional accumulation:
\begin{equation}
\label{equ:res_align}
\left\{
\begin{array}{cc}
\phi_k=\phi_{k-1}\circ\varphi_{k}\\
\varphi_{k}=\mathcal{R}(\phi_{k-1}(\textbf{\textit{I}}^{\rm src}), \textbf{\textit{I}}^{\rm tgt})
\end{array}
\right.
,~~0<k\leq K
\end{equation}
where $\circ$ signifies the composition of two deformation fields, with $\phi_{0}$ being initialized without deformation field. The proposed VMambaMorph network, as illustrated in Fig.~\ref{fig:framework}, integrates a weight-sharing, enhanced fine-grained feature extractor and a U-shaped VMamba-CNN hybrid registration module to accurately estimate the deformation field $\psi_k$, and then to refine the final prediction $\phi_k$. Following the optimal setting validated by \cite{zheng2024residual}, the recursive number is set to ${K}_{\rm t}=2$ in the training process, and varying recursive number ${K}_{\rm i}>1$ is compared in the inference stage to optimize the registration performance.

\subsection{Enhanced Fine-grained Feature Extractor}

The existing methods, such as \cite{balakrishnan2019voxelmorph}, generally combine the two volumes into a unified entry to estimate the voxel-wise spatial correspondences. While this approach may be adequate for registering volumes of the same modality, which naturally exhibit similar visual characteristics, it often proves inadequate for multi-modality registrations characterized by significant differences in appearance. In such cases, it becomes crucial to extract matching features from corresponding anatomical regions across the divergent volumes.

Following MambaMorph \cite{guo2024mambamorph}, a refined yet uncomplicated UNet architecture that features a single down-sampling step for efficient feature extraction. This modification better preserves the local characteristics of images. This, in turn, enhances the detail captured for the task at hand. VMambaMorph leverages a simultaneous learning approach to perform both feature extraction and registration, employing weight-sharing among feature extractors within our proposed model.

\subsection{Training Setting}

The training aim of image registration is to minimize the Loss $\mathcal{L}$ based on two labels $\textbf{\textit{S}}^{\rm tgt}$ and $\textbf{\textit{S}}^{\rm src}$, which is formulated in Equation \ref{loss}.

\begin{equation}
\label{loss}
\min\limits_{\textbf{\textit{w}}} \mathcal{L}_{\rm dice}(\phi(\textbf{\textit{S}}^{\rm tgt}),\textbf{\textit{S}}^{\rm src})+\lambda_{\rm s} \mathcal{L}_{smooth}(\phi)
\end{equation}
where the dense deformation vector field $\phi$ is calculated by: 
\begin{equation}
\phi=\mathcal{R}(\textbf{\textit{F}}^{\rm tgt},\textbf{\textit{F}}^{\rm src};\textbf{\textit{w}})    
\end{equation}
with the feature maps are calculated by:
\begin{equation}    
\textbf{\textit{F}}^{\{\rm tgt,src\}}=\mathcal{F}(\textbf{\textit{I}}^{\{\rm tgt,src\}};\textbf{\textit{w}})
\end{equation}
and $\phi(\cdot)$ denotes the warping operation and is implemented by STN \cite{jaderberg2015spatial}. $\mathcal{L}_{dice}(\cdot,\cdot)$ is the supervised loss based on dice similarity coefficient, and $\mathcal{L}_{smooth}(\cdot)$ denotes the same smooth loss as VoxelMorph \cite{balakrishnan2019voxelmorph,guo2024mambamorph}.

\section{Experiments}

{\bf Implementation Details:} Our code has been developed under Ubuntu 20.04 in Python 3.10.13 using Pytorch 1.13.1 and CUDA 12.1 using a single Nvidia GeForce RTX 3090 GPU, and Intel(R) Intel Core i9-10900K. The runtimes averaged around 15 hours. VMambaMorph and all baseline methods are trained for 300 epoches, the batch size is set to 1, the optimizer is Adam, and the learning rate is initially set to 0.001. The setting is also applied to all other baseline methods directly without any further modification.

{\bf Dataset:} A public benchmark brain MR-CT registration dataset named SR-Reg is utilized in this paper \cite{guo2024mambamorph}. The SR-Reg dataset consists of 180 well-aligned, skull-stripped and intensity-rectified MR-CT pairs, with corresponding segmentation labels, and it is mainly based on the SynthRAD 2023 dataset \cite{thummerer2023synthrad2023,billot2023synthseg,hoopes2022synthstrip}. The original size of each volume is with $192 \times 208 \times 176$ voxels with a resolution of $1 \times 1 \times 1 mm^{3}$. More details can be found at \cite{guo2024mambamorph}. Considering the input size requirement of VMamba, and computational cost on a single GPU, we have resized the volume to $128 \times 128 \times 128$, and randomly select 150 cases for training, 10 cases for validation, and the rest 20 cases for testing.

\begin{table}[ht]
\centering
\caption{Quantitative comparison of VMambaMorph methods against other baseline methods on the SR-Reg testing set.}
\begin{tabular}{l|c|c|c|c||c|c|c}
\hline
\textbf{Model} & \textbf{FeatEx} & \textbf{Dice} & \textbf{HD$_{95}$} & {\bf P}\( _{|J \varnothing |\leq 0} \) & Mem & Par & Time \\ 
&&(\%)&vox&(\%)&(Gb)&(Mb)&(sec)\\
\hline
Initial & - & 62.42\(\pm\)3.29 & 3.73\(\pm\)0.41 & - & -& -& - \\
VoxelMorph \cite{balakrishnan2019voxelmorph} &  \xmark & 71.27\(\pm\)1.95 & 2.11\(\pm\)0.25 & 0.13\(\pm\)0.02 & 1.22 & 0.09 & 0.01  \\
VoxelMorph \cite{balakrishnan2019voxelmorph} &  \cmark & 76.98\(\pm\)1.27 & 1.69\(\pm\)0.14 & {\bf 0.08\(\pm\)0.02} & 1.98 & 0.16 & 0.04\\
TransMorph \cite{chen2022transmorph} &  \xmark & 77.87\(\pm\)1.99 & 1.71\(\pm\)0.19 & 0.66\(\pm\)0.05 & 2.81 & 14.29 & 0.06  \\
TransMorph \cite{chen2022transmorph} & \cmark & 82.31\(\pm\)1.73 & 1.39\(\pm\)0.16 & 0.58\(\pm\)0.04 & 3.27 & 14.57 & 0.08 \\
MambaMorph \cite{guo2024mambamorph} &  \xmark & 79.17\(\pm\)1.78 & 1.60\(\pm\)0.18 & 0.39\(\pm\)0.02 & 2.47 & 7.31 & 0.06 \\
MambaMorph \cite{guo2024mambamorph} & \cmark & 81.92\(\pm\)2.10 & 1.45\(\pm\)0.24 & 0.34\(\pm\)0.00 &3.24 & 7.59 & 0.09\\
\hline
{\bf VMambaMorph} & \cmark & {\bf 82.94\(\pm\)2.01} & {\bf 1.35\(\pm\)0.18} & 1.04\(\pm\)0.05 & 3.25 & 9.64 & 0.10  \\

\hline
\end{tabular}
\label{tab:results}
\end{table}

{\bf Metrics:} We use the the percentage of mean dice-coefficients (Dice), 95\% Hausdorff distance (HD$_{95}$) to measure the registration accuracy, percentage of non-positive Jacobian determinant (P\( _{|J \varnothing |\leq 0} \)) to evaluate the diffeomorphic property of deformation field. The inference time (Time) with the unit of second (s), GPU memory cost (Mem) with the unit of Gigabyte (Gb) and the amount of parameters of network with the unit Mb is used to evaluate the proposed VMamabaMorph with other baseline methods.

{\bf Comparison with State-of-the-Art:} To ensure a fair and thorough evaluation of VMambaMorph, we have selected a range of baseline methods for comparison: the purely CNN-based VoxelMorph \cite{balakrishnan2019voxelmorph}, the Transformer-CNN hybrid TransMorph \cite{chen2022transmorph}, and the recent Mamba-CNN hybrid MambaMorph \cite{guo2024mambamorph}.

\begin{table}[ht]
\centering
\caption{Ablation study of VMambaMorph methods for feature extractor (\textbf{FeatEx}) and recursive training ($K_{\rm t}$) or inference ($K_{\rm i}$).}
\begin{tabular}{l|c|c|c|c|c|c||c|c|c}
\hline
\textbf{Model} & $K_{\rm t}$& $K_{\rm i}$& \textbf{FeatEx} & \textbf{Dice} & \textbf{HD$_{95}$} & {\bf P}\( _{|J \varnothing |\leq 0} \) & Mem & Par & Time \\ 
&&&&(\%)&vox&(\%)&(Gb)&(Mb)&(sec)\\
\hline
Initial & - & -& - & 62.42\(\pm\)3.29 & 3.73\(\pm\)0.41 & - & -& -& - \\
\hline
VMambaMorph & 1 & 1 &  \xmark & 78.46\(\pm\)2.13 & 1.69\(\pm\)0.22 & {\bf 0.33\(\pm\)0.01} & 2.94 & 9.35 & 0.07 \\
VMambaMorph & 1 & 1  & \cmark & 82.49\(\pm\)1.99  & 1.38\(\pm\)0.18 & 0.35\(\pm\)0.01 & 3.25 & 9.64 & 0.10  \\
VMambaMorph & 1 & 2 &  \cmark & 82.84\(\pm\)2.05 & 1.44\(\pm\)0.20 & 1.05\(\pm\)0.01 & 3.93 & 9.64 & 0.20  \\
VMambaMorph & 1 & 3  & \cmark & 82.80\(\pm\)2.06 & 1.47\(\pm\)0.18 & 1.71\(\pm\)0.07 & 3.93 & 9.64 & 0.29 \\
VMambaMorph & 1 & 4  & \cmark & 82.62\(\pm\)2.11 & 1.50\(\pm\)0.18 & 2.29\(\pm\)0.09 & 3.93 & 9.64 & 0.36 \\
\hline
VMambaMorph & 2 & 1 & \cmark & 82.47\(\pm\)2.01 & 1.38\(\pm\)0.18 & 0.35\(\pm\)0.01 & 3.93 & 9.64 & 0.10\\
VMambaMorph & 2 & 2 & \cmark & {\bf 82.94\(\pm\)2.01} & {\bf 1.35\(\pm\)0.18} & 1.04\(\pm\)0.05 & 3.93 & 9.64 & 0.19\\
VMambaMorph & 2 & 3 & \cmark & 82.85\(\pm\)2.07 & 1.47\(\pm\)0.18 & 1.69\(\pm\)0.06 &  3.93 & 9.64 & 0.28 \\
VMambaMorph & 2 & 4 & \cmark &82.65\(\pm\)2.16 & 1.51\(\pm\)0.17 & 2.26\(\pm\)0.08 & 3.93 & 9.64 & 0.40  \\
\hline
\end{tabular}
\label{tab:ablation}
\end{table}

\begin{figure*}[ht!]  
\centering  
\includegraphics[width=\linewidth]{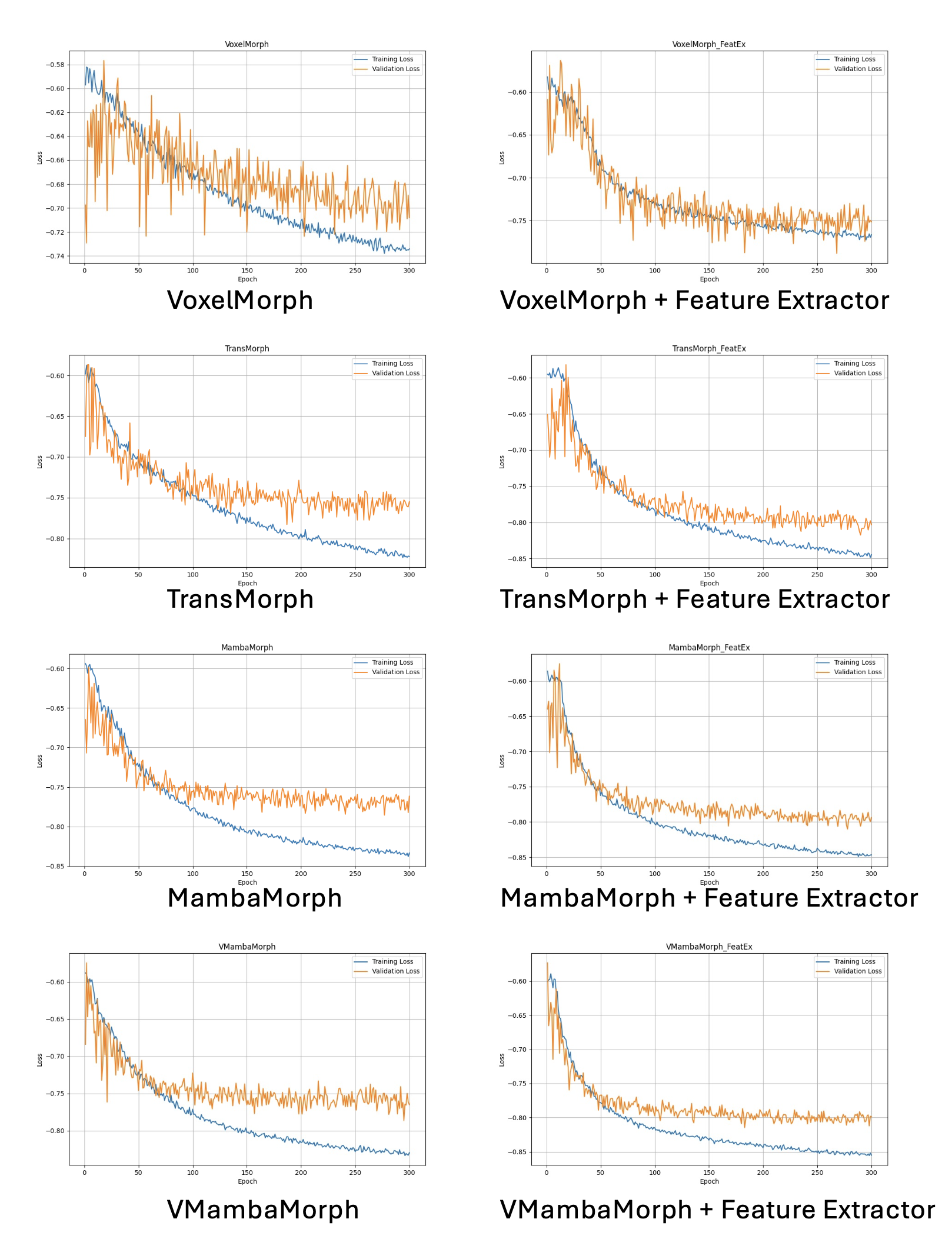}  
\caption{The Training History of VoxelMorph, TransMorph, MambaMorph, and VMambaMorph with and without the Feature Extractors.}  
\label{fig:history}  
\end{figure*}

{\bf Training History:} Figure \ref{fig:history} depicts the training progression for all image registration methods, with the X-axis representing the training epoch and the Y-axis denoting the loss observed on the training and validation sets. It is important to highlight that the incorporation of the fine-grained feature extractor substantially reduces and stabilizes the training loss. Notably, VMambaMorph demonstrates the most consistent performance, maintaining the lowest level of loss on the validation set throughout the training process. This evidence underscores the efficacy of VMambaMorph in learning from the data.

{\bf Results:} The quantitative results from a direct comparison of VMambaMorph with other baseline methods across all evaluation metrics are reported in Table~\ref{tab:results}. Our proposed method and the top-performing results are highlighted in {\bf bold}. It is noteworthy that VMambaMorph achieves 16\%, 6\%, and 1\% higher Dice scores compared with VoxelMorph, TransMorph, and MambaMorph, respectively.

{\bf Ablation Study:} To validate the effectiveness of the fine-grained feature extractor, we conducted an ablation study. The performance impact of the feature extractor (denoted as {\bf FeatEx}) is evaluated in conjunction with all methods in Table~\ref{tab:ablation}. The deployment of {\bf FeatEx} is indicated with a \cmark, while its absence is marked with an \xmark. All methods demonstrate improvement when the feature extractor is utilized. To validate the proposed recursive registration framework, the bottom of Table~\ref{tab:ablation} further explores the affect of varying recursive numbers in training and inference.

\section{Conclusion}

In this paper, we introduce VMambaMorph, a multi-modality deformable image registration framework based on the Visual State Space Model. The architecture of VMambaMorph is developed with a hybrid registration module with 3D VMamba-based network blocks and 3D CNN-based network blocks. Following the recent advances in feature learning for multi-modality registration, VMambaMorph incorporates the recursive registration framework with a pair of fine-grained feature extractor to achieve high-level spatial features more effectively. Our validation on a public MR-CT registration dataset indicates that VMambaMorph surpasses recent methodologies such as MambaMorph, as well as established techniques like VoxelMorph and TransMorph. This paves the way for future applications of VMamba in broader computer vision tasks.

\bibliographystyle{splncs04}
\bibliography{mybibliography}

\end{document}